\documentclass[11pt,a4paper]{article}
\usepackage{times,latexsym}
\usepackage{url}
\usepackage[T1]{fontenc}

\usepackage[acceptedWithA]{tacl2018v2}

\usepackage{xspace,mfirstuc,tabulary}

\newif\iftaclinstructions
\taclinstructionsfalse 
\iftaclinstructions
\renewcommand{\confidential}{}
\renewcommand{\anonsubtext}{(No author info supplied here, for consistency with
TACL-submission anonymization requirements)}
\newcommand{\instr}
\fi

\iftaclpubformat 

\else

\fi

\usepackage{graphicx}
\graphicspath{ {./images/} }

\usepackage{amsmath} 
\usepackage{amsfonts}

\usepackage{array}

\makeatletter
\newcommand*{\centerfloat}{%
  \parindent \z@
  \leftskip \z@ \@plus 1fil \@minus \textwidth
  \rightskip\leftskip
  \parfillskip \z@skip}
\makeatother

\usepackage{wrapfig}

\usepackage[caption=false]{subfig}

\usepackage{floatrow}
\newfloatcommand{capbtabbox}{table}[][\FBwidth]

\newcommand{\subsubsubsection}[1]{\paragraph{#1}\mbox{}\\}
\title{On the Use of Linguistic Features for the Evaluation of Generative Dialogue Systems}

\author{
 Ian Berlot-Attwell $^{1,2}$, Frank Rudzicz $^{1,2,3}$ \\
$^1$ Department of Computr Science, University of Toronto; \\
$^2$ Vector Institute for Artificial Intelligence;\\
$^3$ Unity Health Toronto\\
  \small{\texttt{\{ianberlot, frank\}@cs.toronto.edu}}\\  \\
}

\date{}

\begin{document}
\maketitle
\begin{abstract}
  Automatically evaluating text-based, non-task-oriented dialogue systems (i.e., `chatbots') remains an open problem. Previous approaches have suffered challenges ranging from poor correlation with human judgment to poor generalization and have often required a gold standard reference for comparison or human-annotated data. 
  
   Extending existing evaluation methods, we propose that a metric based on linguistic features may be able to maintain good correlation with human judgment and be interpretable, without requiring a gold-standard reference or human-annotated data. To support this proposition, we measure and analyze various linguistic features on dialogues produced by multiple dialogue models. We find that the features' behaviour is consistent with the known properties of the models tested, and is similar across domains. We also demonstrate that this approach exhibits promising properties such as zero-shot generalization to new domains on the related task of evaluating response relevance.
\end{abstract}

\section{Introduction}

Automatic evaluation of generative non-task oriented dialogue systems remains an open problem. Given a context $c$ (e.g., ``Speaker 1: Hello, how are you? Speaker 2: I'm fine, yourself?") these systems produce a response $r$ (e.g., ``Speaker 1: Pretty good, all in all"). Two of the major difficulties in evaluating such a system are the inherently qualitative nature of a response's quality, and the vast space of acceptable responses. A variety of approaches have been tried, but all suffer from problems such as dependence on human-annotated data, poor generalization between domains, requiring human gold references, overfitting to a specific dialogue model, and so on (a full breakdown can be found in Section \ref{prevwork}).

In this work, we propose that an approach based on linguistic features can 1) be implemented without any human-annotated data or test-time gold references, 2) be easily interpreted, and 3) exhibit zero-shot generalization to new domains. To support our claims, we evaluate and analyze a set of linguistic features on dialogues from different domains produced by a variety of dialogue systems. Furthermore, we train an unsupervised feature-based model for the related task of relevance-prediction and demonstrate superior performance to our unsupervised baselines as well as zero-short generalization to other domains.

\section{Related Work}\label{prevwork}

Historically, non-task-oriented dialogue system evaluation has been approached by viewing dialogue as a machine-translation problem with the context as the source language and the response as the translation. Under this formulation, machine translation metrics based on $n$-gram overlap (e.g., BLEU, METEOR) or word embedding similarity were used to evaluate system responses, based on human references. However, these approaches correlated very poorly with human judgment \cite{HowNotTo}. 

This incongruity can be explained by several observations. First, the majority of these translation metrics were designed to have multiple gold-standard translations, whereas dialogue corpora typically have only one human response per context. Second, these metrics only compare the system's response against the gold-standard response -- they fail to consider the context that generated these responses.

To some extent, $n$-gram models are disadvantaged in that many valid dialogue responses lack word overlap of any kind. Similarly, approaches based on word embeddings are disadvantaged in that semantically similar words can still create very inappropriate responses \cite{HowNotTo}.

\subsection{BLEU-Inspired Approaches}

Despite these problems, there have been attempts to modify BLEU for dialogue. One such attempt is $\Delta$BLEU \cite{deltaBLEU}; in this approach, a subset of a pre-existing dialogue corpus is extended with additional possible responses by automatically finding $\langle context,response\rangle$ pairs on the full corpus similar to those in the subset, and adding the responses to the subset. Each reference is scored by humans on a 5-point Likert scale, and then re-scaled to $\left[-1,1\right]$. These reference scores are then used in a modified BLEU -- specifically the brevity penalty is unchanged, and the corpus-level $n$-gram precision is re-defined as:

\begin{equation*}
   \frac{ \displaystyle \sum_i \sum_{g \in ngrams(h_i)} \max_{j:g\in r_{ij}} \{w_{ij} \min (\#_g(h_i), \#_g(r_{ij})) \} }{\displaystyle \sum_i \sum_{g \in ngrams(h_i)} \max_{j} \{w_{ij} \#_g(h_i)\} },
\end{equation*}
where $h_i$ is the hypothesis for some context $c_i$, $r_{ij}$ and $w_{ij}$ are a reference and its weight for context $c_i$, and $\#_g(x)$ is the count of the $n$-gram $g$ in the text $x$ \cite{deltaBLEU}. Despite achieving a Spearman's correlation of $0.485$ on groups of 100 responses (compared with BLEU's performance with a single response of $0.26$), the sentence-level correlation was still poor (less than $0.1$). Another work loosely based on this approach achieved a sentence-wise Pearson's correlation of $0.514$ \cite{deltaBLEU}. The approach first hired humans to create new responses for every context  under zero or one constraints (e.g., the response must be less than $X$ characters, or only $X\%$ of the context is visible). Next, these responses were human-scored based on pairwise-win rate (i.e., the percentage of comparisons with which it is preferred over another possible response). Finally, they were sorted into decreasing order of win rate, and used to train an SVR to predict the pairwise-win rate of a new response from the $\left[-1,1\right]$-normalized word error rate (WER) of the new response with the top $M$ human-created responses \cite{deltaBLEU}. However, both of these approaches have the problem that, for each context, a set of human-scored possible responses is required. As such, these approaches become impractical as the set of contexts expands.

\subsection{Surrogate Tasks}

Another approach for dialogue evaluation has been to evaluate performance on the surrogate task of `next utterance classification' (NUC) \cite{NUC}. NUC is the task of selecting the most appropriate response from a list of multiple possible responses, and is evaluated using Recall@$k$ (i.e., the percentage with which the correct response is among the top $k$ responses suggested by the system). The performance of dialogue systems on this task is still inferior to human experts, and thus it is a useful but insufficient metric for dialogue system evaluation \cite{NUC}. A drawback of NUC is that it may be impossible or difficult to compute Recall@$k$ depending on the model. For instance, TF-IDF retrieval models (see Section \ref{describe_tf-idf_section}) cannot be evaluated with NUC unless all possible responses are in its dataset, and it is often intractable to calculate the probability of a response under a latent variable model such as VHRED, (see Section \ref{describe_VHRED_section}).

\subsection{Regression-Based Approaches}

\citet{AutoTuringTest} worked to solve this problem by instead formulating the dialogue evaluation task as a regression problem. Their approach was to construct a dialogue corpus where each response was annotated with an average human response on a rating scale of the perceived quality. This corpus was then used as training data for a neural net. The net (ADEM) functioned by first encoding the context, system response, and gold standard response as vectors ($\mathbf{c}$, $\mathbf{\hat{r}}$, and $\mathbf{r}$, respectively) using a hierarchical RNN. The score was then computed using learned matrices $M$ and $N$ as: $(\mathbf{c}^TM\mathbf{\hat{r}} + \mathbf{r}^TM\mathbf{\hat{r}} - \alpha)/\beta$ (the values $\alpha, \beta \in \mathbb{R}$ being constants chosen at the start of training so that initially  all predictions are in the range $[1,5]$). This approach yielded superior results than re-purposed MT methods, achieving an utterance-level Pearson's correlation of $0.436$ on the test data (for comparison, the best of the other methods was ROUGE, with a Pearson's correlation of $0.147$).

However, this approach is not without its own flaws. ADEM has had difficulty transferring to other datasets -- a behaviour that is currently believed to originate in insufficient training data possibly combined with an inherent bias in the dataset \cite{RyanLowe}.  This is a problem inherent to all trained black-box evaluators, which have the potential to overfit to the training data, the domain of the data, or even the methods used to generate the subpar examples in the training data.  This vulnerability to overfitting was shown by \citet{ReEvalADEM} who demonstrated that ADEM can be tricked by methods as simple as reversing or jumbling word order.

More recently, \citet{deriu-cieliebak-2019-towards} developed AutoJudge, a spiritual successor to ADEM that does not require a gold-standard response at evaluation time. Although it achieved strong correlation of $0.577$, it is once again dependent on a dataset of human-annotated dialogues. Furthermore, they found that their trained metric suffered from instability due to the relatively small size of their created dataset and, more importantly, they could not use AutoJudge as a reward signal for reinforcement learning. The authors concluded that although AutoJudge can distinguish the good and bad utterances of trained systems, there are classes of bad responses that trick AutoJudge. They suggested that modifying the training procedure would be required for AutoJudge to be able to identify these, leading us to conclude that AutoJudge is overfitting to its training data. We therefore also speculate that AutoJudge may share ADEM's tendency to overfit to its domain.

\subsection{Adversarial Approaches}

Adversarial evaluation also suffers from the same potential overfitting problems, but does not require human-annotated quality data  \cite{AdvEvDialogue}. In this approach, a neural net is trained as a binary classifier that discriminates between human and machine-generated responses given the context. In an experiment to discriminate between human and RNN-generated dialogue, the experimenters achieved an accuracy of 62.5\%, and found the discriminator correlated strongly with length and weakly with the generator likelihood (it has been long observed that generated responses tend to be shorter, and that highly likely responses tend to be generic) \cite{AdvEvDialogue}. However, as noted by the authors, there is no evidence that a dialogue system that fools the discriminator would necessarily have superior human evaluation \cite{AdvEvDialogue}.

\section{Methods}\label{DialogueModUsed}

We propose that, through linguistic features, a metric for generative dialogue systems can be created that is interpretable, does not require gold-standard responses, does not require human annotated data, and generalizes to other dialogue domains. As a proof-of-concept, we perform two experiments to demonstrate that a feature-based approach can display these properties.

First, we generate responses to the same contexts using the dialogue models described in this section and analyze the distribution of features seen over different sources. We expect different sources to exhibit different distributions based on the properties of the generative model. 

Second, we empirically demonstrate that a feature-based approach can demonstrate domain generalization, and be produced and evaluated in an unsupervised manner. Ideally, we would want to measure the correlation of our metric with human ratings for the quality of the response. Unfortunately, to our knowledge, there is at present no publicly available dataset of dialogues annotated directly with response quality. Instead, we use the HUMOD dataset \cite{Merdivan_2020} and infer the relevance of responses to contexts rather than the quality. We then compare performance to supervised baselines and measure model performance when fitted on one domain, and tested on another. Response relevance is clearly not a perfect proxy for response quality, but our goal is to show the feasibility of the approach rather than to present a final solution, and it is self-evident that response relevance is a major component of overall response quality. To demonstrate that a metric can be produced from only sample conversations, we train an unsupervised logistic regression model to predict response relevance and measure correlation to human raters on a different domain. 

\subsection{Generative dialogue models}

\subsubsection{Simple baselines}

We have three simple baselines. The simplest, {\bf Collapsed}, returns ``{\em I don't know}" to all inputs; this is representative of a model that has undergone mode collapse, giving a highly generic response to all contexts \cite{generic_resp1, HRED}. The second baseline, {\bf Random}, returns a random response from the training set, representing a model that yields very natural text but which is independent of the context. The final baseline, {\bf Gold Reference}, is the true human response from the corpus.

\subsubsection{TF-IDF retrieval-based model}\label{describe_tf-idf_section}

As described by \citet{UbuntuCorpus}, this retrieval model works by computing the TF-IDF vector for each context in the training data. When a new context is provided, its TF-IDF vector is computed, the closest context in the training data is found using cosine similarity and, finally, the response corresponding to the closest context is returned. Note that we used TF-IDF vectors as we wanted to include a non-neural dialogue model in our analysis.

\subsubsection{Generative models}\label{sectionGenModels}

\subsubsubsection{LSTM RNNLM}

The simplest generative dialogue model used in our analysis is a RNNLM using 2000 LSTM hidden units which was pretrained by \citet{VHRED}. 

\subsubsubsection{HRED}

The second model is a hierarchical recurrent encoder-decoder. The HRED views dialogue as being divided into utterances $\mathbf{U_1}, \mathbf{U_2}, \dots, \mathbf{U_M}$ alternating between two interlocutors. Each utterance decomposes into tokens such that $\mathbf{U_i} = (w^{(i)}_1, w^{(i)}_2, \dots, w^{(i)}_{N_i})$. The HRED uses three RNNs to generate the dialogue. The first RNN independently encodes utterances to a vector representation. The second (the context RNN) encodes the sequence of utterance vectors into the context representation. The final RNN is the decoder, which is conditioned at all time steps on the context vector \cite{HRED}.

The pretrained model contains 500 GRU hidden units in the encoder RNN, 1000 GRU hidden units in the context RNN, and 500 GRU hidden units in the decoder RNN \cite{VHRED}.

\subsubsubsection{VHRED}\label{describe_VHRED_section}

The final generative architecture is a variational hierarchical encoder-decoder (VHRED) \cite{VHRED}. The VHRED maintains the overall design of the HRED, except that the decoder is conditioned on the concatenation of the context vector with a latent variable that is sampled once per decoded utterance. The latent variable follows a multivariate normal distribution with diagonal covariance, where the exact parameters are a learned function of the context vector. 

\subsection{Relevance prediction model}\label{ulrof}

To observe the properties of a feature-based approach, we train a model to predict response relevance given the context. Specifically, we perform unsupervised training of a logistic regression model, where $0$ corresponds to maximum relevance and $1$ to minimal relevance. For a context $c$ and response $r$, the model is defined below where $\mathbf{f}(c,r)$ is a vector of computed linguistic features, $\mathbf{w}$ are the learned feature weights,  $b$ is a learned bias, $\sigma$ is the sigmoid function, and $y(c,r)$ is the predicted relevance: 

$$ y(c,r) = \sigma \left(\mathbf{w}^T\mathbf{f}(c,r) + b\right) $$

To train the model, we developed a loss function inspired by triplet loss \citep{triplet}. Specifically, for context $c$, true human response $r$, randomly sampled human response $r'$, and margin $m$ (a hyperparameter) we use the loss in Equation \ref{loss2}. Note that Equation \ref{loss1} is triplet loss modified to use the model prediction as a measure of distance as opposed to the norm of two embeddings, and Equation \ref{loss2} is a function of the triplet loss specifically for logistic regression to prevent the sigmoid in the model from driving the gradient to zero. 

\begin{equation}\label{loss1}
    \mathcal{L}_{t}(c,r) =  \max \left( y(c,r) - y(c,r') + m, 0 \right)
\end{equation}

\begin{equation}\label{loss2}
    \mathcal{L} (c,r) = -\log \left(1 + m - \mathcal{L}_{t}(c,r) \right)
\end{equation}

Note that as we trained this model to predict response relevance instead of overall quality we only use a subset of our linguistic features in our experiments.

\section{Linguistic features}

We evaluate the following linguistic features on the responses generated by the models outlined in Section \ref{sectionGenModels}.

\subsection{Acknowledgement (\texttt{Ack})}
To measure the degree to which the response addresses the context, the contentful words of the response are extracted using SpaCy (i.e., nouns, verbs, adverbs and adjectives that are not stop words) and we use WordNet \cite{WordNet} to determine the percentage of these contentful words which have a synonym within the context. If the response contains no contentful parts of speech, then NaN is returned. When training or evaluating our unsupervised model, we replace NaN with $0$.

\subsection{Relatedness (\texttt{Rel})}\label{section_relatedness}

To measure the semantic relatedness of the contentful words that do not have a synonym in the context, we use GloVe vectors pretrained on the Twitter Corpus \cite{glove} and determine the average cosine distance between the contentful word and the most similar word in the context. Words for which we do not have a word vector are ignored. We empirically experiment with 25- and 200-dimensional GloVe vectors. 

The motivation for looking at contentful tokens which do not have a synonym in the context is that these tokens provide  new information to the response. Meaningful dialogue responses should add new related information. For example, given the context ``{\em So, any hobbies? It's soccer and badminton with me.}", the answer ``{\em My hobbies are my hobbies}" fails to add information, the answer ``{\em Asphalt}" fails to add relevant information, whereas the response ``{\em Yodelling}" adds new relevant information. Note that in the context, the only token that is truly related to ``{\em Yodelling}" is ``{\em hobbies}"; soccer and badminton and very different concepts to yodelling but become part of a coherent whole through the link of hobbies. To allow for examples such as this, in computing \texttt{rel} we only consider the most similar word in the context.  Thus, this metric would ideally measure the relatedness of the new information in the response with the context. Consequently, \texttt{Rel} is defined to be zero if there are no contentful words without a synonym in the context.

\subsection{N-gram precision}

As a baseline measure of coherence and similarity, BLEU's clipped $n$-gram precision \cite{bleu} is calculated between the stemmed response and the stemmed context for 2-, 3-, and 4-grams. Stemming is performed using NLTK \cite{nltk}.

Note that this feature is very different to BLEU.  First, BLEU is calculated between the generated response and the ``true" response (i.e., the gold reference) whereas this feature is calculated between the generated response and the context. This makes the feature context-dependent unlike BLEU, and removes the need for a ``true" human response which may not be available. Second, unlike BLEU we stem the context and response before calculating the precision, this helps to account for changes in verb tense or number between the response and context. Finally, unlike BLEU we do not apply the brevity penalty, nor average the $n$-gram precision over several values of $n$.

\subsection{Normalized language tool (\texttt{LTNorm})}

To measure grammaticality, we calculate the number of LanguageTool errors detected, normalized by response length, specifically $1 - \frac{\# errors}{\# tokens}$ as described by \citet{GEC_referenceless}. To ensure the focus on grammaticality only Grammar, Collocation, and Capitalization errors are considered. This also reduces spurious errors such as false positive spelling errors for technical terms in the Ubuntu corpus.  

\subsection{Neural acceptability (\texttt{nn\_acc})}

 \citet{nnacc} presented a measure of acceptability using a neural classifier to label sentences as acceptable or unacceptable, trained on a developed corpus.  The acceptability of each response is calculated using the pretrained model from \citet{nnacc}.

Note that \texttt{LTNorm} measures (syntactic) grammaticality whereas \texttt{nn\_acc} is meant to measure acceptability, according to a native speaker. This distinction can be seen in garden-path sentences that are considered grammatical, but unacceptable. For instance, ``{\em the old man the boat}" actually means that ``{\em the old [people] staff the boat}".

Note that unlike the other features, \texttt{nn\_acc} is not interpretable -- we do not understand how the feature is calculated. However, the ensemble of features can still be interpretable as we know the aspect it should be capturing. For an ensemble such as logistic regression we can trace its decisions to a set of features that are abnormally low or high. If, for instance, \texttt{nn\_acc} is abnormally low then we know that our model is scoring poorly as the metric believes that the model has difficulties producing acceptable responses. From here it is a straightforwards matter to determine if the problem lies with the model, or the metric.

\section{Data}

We use two corpora for analysis and unsupervised training  -- the subsets of the Ubuntu Corpus \cite{UbuntuCorpus} and Twitter Dialogue Corpus \citep{TwitterCorpus} used by \citet{VHRED} to train their VHRED, HRED, and LSTM generative dialogue models. Dialogue is divided into contexts and responses, where the contexts contain multiple turns (i.e., continuous sections of text by a single person) and the response is a single turn. Turns are subdivided into utterances -- complete sentences or consecutive posts by the same speaker. The Twitter Dialogue corpus consists of dialogues from Twitter, and the Ubuntu Corpus from conversations from the Ubuntu technical support chat.

For evaluation we use the HUMOD Dataset \citep{Merdivan_2020} consisting of dialogues extracted from the Cornell movie dialogues dataset \citep{Danescu-Niculescu-Mizil+Lee:11a}. Each dialogue is annotated with three human ratings for the relevance of the true response, and three human ratings for the relevance of a response selected from the dataset at random. Ratings are on a Likert scale from $1$ to $5$, with $5$ being the most relevant. Following \citep{Merdivan_2020}, we split off a test set of $1000$ dialogues. As their exact train/test split is not provided, we take the first $7500$ dialogues as training data for supervised models, the following $1000$ dialogues as a validation set, and the last $1000$ dialogues as a test set.

\subsection{Pre- and post-processing}
The raw subset of the Ubuntu Corpus comes with modest preprocessing (e.g., simple word tokenization, end-of-utterance and -turn tags, and the replacement of unknown words in the response with \texttt{**unknown**}). We performed preprocessing on the Twitter Corpus to replace URLs and references with fixed symbols, to remove emoticons, and to apply NLTK's TweetTokenizer \citep{nltk}. We perform no further preprocessing for the retrieval models. 

All of the contexts and responses produced by the various systems are then further postprocessed before the linguistic features are calculated. Specifically, we remove the end-of-utterance tags and detokenize grammatical symbols and possesive ``'s" (e.g., `{\em `Bob 's ten . }" $\rightarrow$ ``{\em Bob's ten.}") to reduce noise in the grammaticality metrics. For the Twitter Corpus, all responses are also converted to lower case.

As for the HUMOD dataset, it comes with no preprocessing and we performed none.

\section{Results}

\subsection{Details by feature}

We begin with our analysis of the features. Our primary observation is qualitative -- the distribution of features across the Ubuntu and Twitter corpora tend to be similar, indicating that a feature-based approach may generalize well. In the following sections, we go into more detail and note where each feature did or did not meet our expectations based on the properties of the generative models used.

Specifically, we produce boxplots of the various distributions and perform the paired sign test between the features of each dialogue model and the gold standard (the true human response to the context). Pairing is by context; the null hypothesis of the test is that the median of the difference between the values is zero. We use a significance level of $0.05$; as we perform 60 tests (tests for 6 dialogue models and 5 features over 2 domains) we apply Bonferroni correction and round down our significance threshold to 8.3e-4.

\subsubsection{Acknowledgement}

\begin{figure*}[ht!]
\centerfloat
\includegraphics[width=0.75\linewidth]{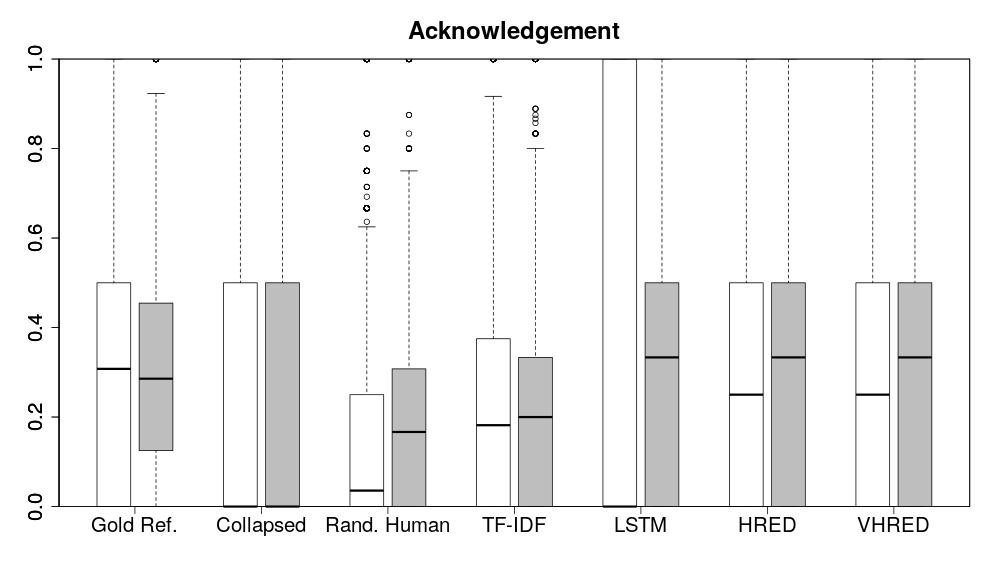}
\caption{Boxplot of Acknowledgement on the subset of data where defined. Results for the Ubuntu Corpus are plotted in white and results for the Twitter Corpus are plotted in grey.}
\label{box_ack_cont}
\end{figure*}

\begin{table}[ht]
\centering
\begin{tabular}{lrrrr}
  \hline
Model & $p_\text{Ub}$ & $\mu_\text{Ub}$ & $p_\text{Tw}$ & $\mu_\text{Tw}$ \\ 
  \hline
Gold Ref & - & 0.3280 & - & 0.3029 \\ 
  Collapsed & 4e-14* & 0.3077 & 2e-16* & 0.1922 \\ 
  Random & 6e-16* & 0.1506 & 2e-16* & 0.2009\\ 
  TF-IDF & 6e-16* & 0.2348 & 4e-16* & 0.2341 \\ 
  LSTM & 4e-16* & 0.3490 & 5e-05* & 0.3327\\ 
  HRED & 1.6e-03 & 0.3283 & 5.6e-02 & 0.3057 \\ 
  VHRED & 3.4e-02 & 0.3324 & 7.4e-01 & 0.3175 \\ 
   \hline
\end{tabular}
\caption{Paired sign test on \texttt{Ack} values between the Gold Reference and all other categories. Data points paired by the context that elicited the response. The p-values on the Ubuntu and Twitter datasets are under reported under the columns $p_\text{Ub}$ and $p_\text{Tw}$ respectively. Stars indicate statistical significance at the Bonferroni-corrected $0.05$ level.  Mean \texttt{Ack} for the Ubuntu and Twitter corpora are also reported under $\mu_\text{Ub}$ and $\mu_\text{Tw}$ respectively.} 
\label{Ack-sign-table}
\end{table}

As \texttt{Ack} may be NaN, we only analyze \texttt{Ack} on a subset of the data where \texttt{Ack} is defined. For the paired sign test, we only considered pairs where \texttt{Ack} was non-NaN for both models. We found that \texttt{Ack} was defined on $100\%$ of the Twitter data, and $88.86\%$ of the Ubuntu data. We name this subset `the contentful set'. The results on the contentful set are plotted in Figure \ref{box_ack_cont}, the results of the sign-test and the mean for each response source are in Table \ref{Ack-sign-table}. As expected, the Collapsed and Random baselines have low medians. Note that VHRED and HRED have medians that are not significantly different from that of the Gold Reference; as expected, the most advanced generative models are the most similar. We also see that, with the exception of LSTM, all sources have similar distributions across the two different domains.

\subsubsection{Relatedness}

\begin{figure*}[ht]
\centerfloat
\includegraphics[width=0.75\linewidth]{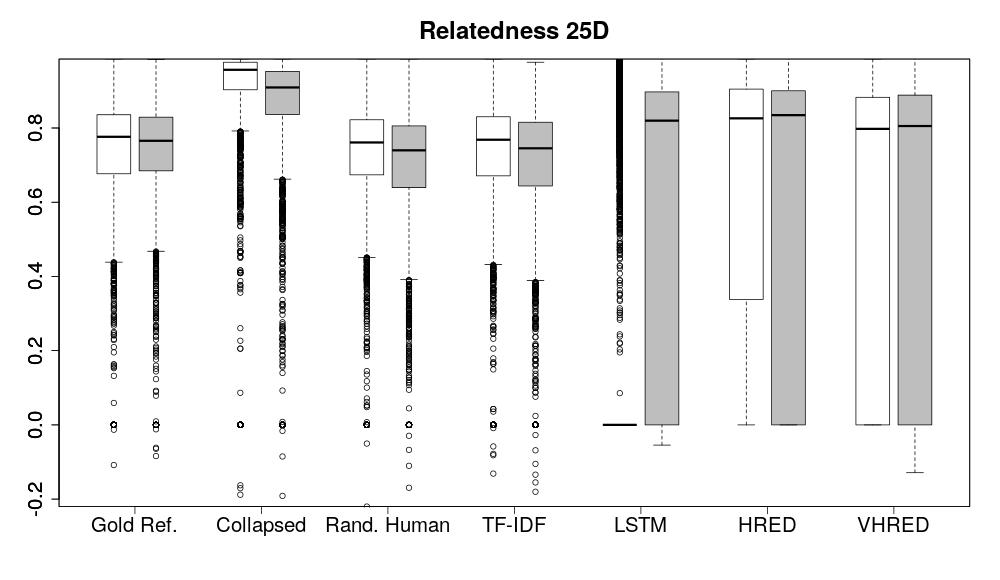}
\caption{Boxplot of relatedness of contentful words that do not have synonyms in the context, specifically using 25D GLoVe vectors. White boxplots are of the Ubuntu dataset, and grey boxplots are of the Twitter dataset. The distribution of relatedness of the gold standard is very similar to that of the Random Human, suggesting that we are failing to measure the semantic relatedness of the new information in the response with the context. The Collapsed distribution also highly corroborates this conclusion.}
\label{box_rel25}
\end{figure*}

Unfortunately, this feature did not perform as well. From Figure \ref{box_rel25}, we can see that the distribution of relatedness of the gold standard is very similar to that of the Random baseline on both the Twitter and Ubuntu corpora, suggesting that we are failing to measure the semantic relatedness of the new information in the response to the context. We attempted to use higher dimensional word embeddings (200D) as well as training GloVe embeddings on Ubuntu data, but the undesired behaviour persisted. Interestingly enough, this feature does seem to contain some useful information, as its inclusion improves performance in the relevance prediction task.

\subsubsection{N-gram reuse}

\begin{figure*}[ht]
\centering
\includegraphics[width=0.75\linewidth]{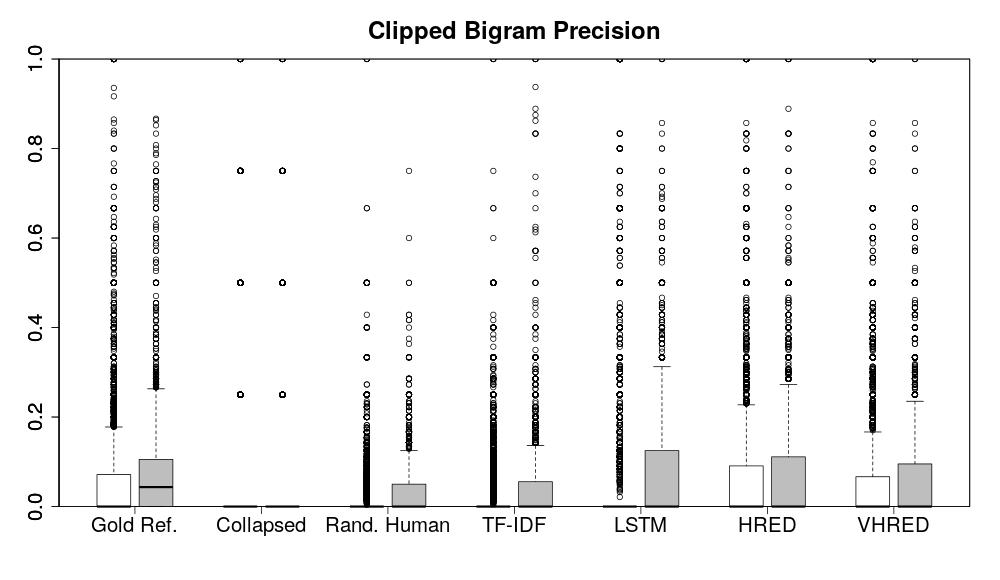}
\caption{Boxplot of clipped bigram reuse between stemmed response and stemmed context. White boxplots are of the Ubuntu dataset, grey boxplots are of the Twitter dataset.}
\label{box_2gram}
\end{figure*}

\begin{table}[ht]
\centering
\begin{tabular}{lrrrr}
  \hline
Model & $p_\text{Ub}$ & $\mu_\text{Ub}$ & $p_\text{Tw}$ & $\mu_\text{Tw}$ \\ 
  \hline
Gold Ref & - & 0.0554 & - & 0.0741 \\  
  Collapsed & 2e-16* & 0.0697 & 4e-16* & 0.0329 \\ 
  Random & 2e-16* & 0.0125 & 4e-16* & 0.0319 \\
  TF-IDF& < 1e-16* & 0.0217 & 2e-16* & 0.0390 \\
  LSTM & 2e-16* & 0.0409 & 2.1e-04* & 0.0778 \\ 
  HRED & 7.1e-01 & 0.0650  & 1e-12* & 0.0691 \\  
  VHRED & 2e-16* & 0.0564 & 4e-16* & 0.0613 \\ 
   \hline
\end{tabular}
\caption{Paired sign test on \texttt{2gram} values between the Gold Reference and all other categories.} 
\label{2gram-sign-table}
\end{table}

Figure \ref{box_2gram} shows that the gold standard, HRED, and VHRED are the only models with considerable bigram reuse. As expected, the Collapsed and Random baselines have negligible $n$-gram reuse. In fact, Random has the lowest average bigram reuse of all the sources. Interestingly, on the Ubuntu data, the LSTM and TF-IDF models also have quite low bigram reuse. For TF-IDF, this may be caused by a combination of TF-IDF vectors ignoring word order and data sparsity. For the LSTM, the most likely explanation is the well documented tendency of the model to produce short and generic responses. In the Ubuntu data, approximately 64\% of the LSTM responses are just 1- or 2-token responses. As expected, Random has the lowest bigram precision.

Note that, for the Twitter dataset, there is some inflation of bigram reuse as all tweets start with a symbol referring to a speaker and the @ sign.

Unfortunately, trigram and four-gram reuse are less useful, as over 75\% of all samples have zero reuse. We also observe that the VHRED model consistently has a lower average $n$-gram reuse than the HRED model. This is to be expected, as VHRED has additional stochasticity in the form of a latent variable. As such, it is reasonable to believe that VHRED would be more likely to attempt trajectories in the response that do not reuse $n$-grams from the context. Unexpectedly, as seen in Table \ref{2gram-sign-table}, HRED is the only model with a median that is not significantly different from the gold reference, occurring only on the Ubuntu data (on the Twitter data it has the largest $p$-value, but is still significantly different).

\subsubsection{Language tool errors \& neural acceptability}

\begin{figure*}[ht]
\centerfloat
\includegraphics[width=0.7\linewidth]{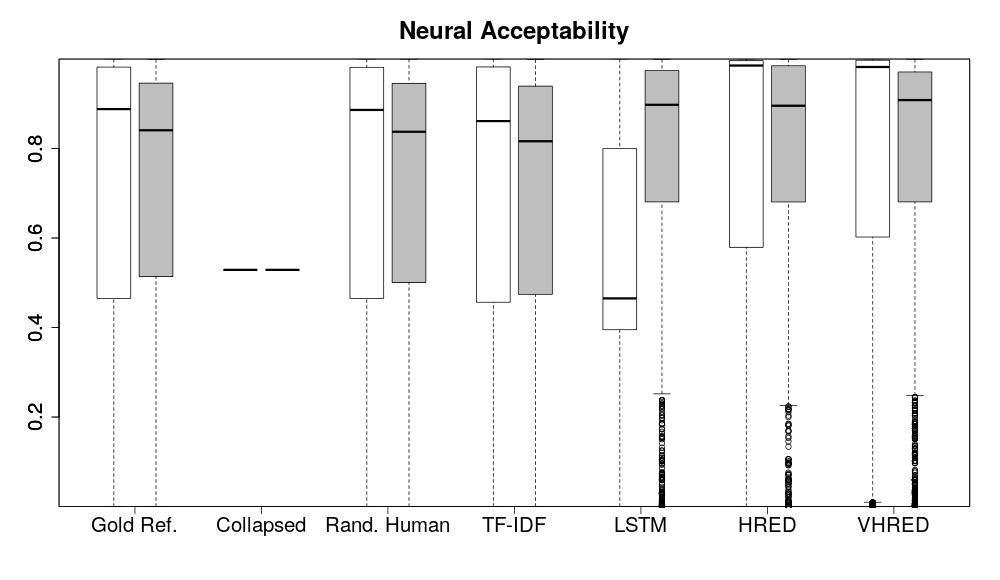}
\caption{Boxplots of Neural Acceptability. White boxes are boxplot over Ubuntu dataset, and grey boxes are over the Twitter dataset.} \label{fig:rec_25}
\end{figure*}

As expected for normalized language tool errors, the median of the random human responses is not significantly different from that of gold reference, and the TF-IDF model is only just significantly different. Similarly, neither TF-IDF nor Random have significantly different medians for Neural Acceptability. We can see similar distribution in neural acceptability regardless of domain in Figure \ref{fig:rec_25}.

\begin{table}[hb!]
    \centering
    \begin{tabular}{lrr}
        \hline
       Unsupervised Model & \multicolumn{1}{c}{r} & p-value \\
       \hline
       BLEU-4 &	0.055 & 0.08 \\
       ROUGE & $-$0.035 & 0.26 \\
       METEOR & $-$0.017 & 0.59 \\
       ULRoF-1 on HUMOD & 0.312 & 6.27e-24\\
       ULRoF-1 on Twitter & 0.283 & 9.63e-20  \\
       ULRoF-1 on Ubuntu & 0.292 & 7.58e-21 \\
       ULRoF-2 on HUMOD & \textbf{0.323} & 2.58e-24 \\
       ULRoF-2 on Twitter & 0.259 & 1.67e-14 \\
       ULRoF-2 on Ubuntu & 0.303 & 6.41e-21 \\
       \hline
    \end{tabular}
    \caption{Comparison of our model's Pearson's correlation with human raters to that of all unsupervised baselines reported by \citet{Merdivan_2020}. Observe that we have significant correlation even when the model has not seen the domain before test-time, and has been trained on the considerably different domain of tech-support chat logs. Note that as human raters give higher scores for more relevant responses and ours does the opposite, we negate our model's relevance score before calculating the correlation.}
    \label{tab:ulr_perf}
\end{table}

\subsection{Unsupervised relevance prediction}

For our unsupervised relevance prediction experiments, we train our model from Section \ref{ulrof} over 20 epochs using ADAM \citep{ADAM} and a learning rate of $0.1$. The performance of our unsupervised model, as compared to all unsupervised baselines reported by \citet{Merdivan_2020}, is in Table \ref{tab:ulr_perf}. Note that we tried two variants of our Unsupervised Logistic Regression of Features (ULRoF) model, ULRoF-1, which only uses \texttt{Ack} as well as the n-gram precisions, and ULRoF-2, which also uses \texttt{Rel} (note that we computed two \texttt{Rel} values, with the 25D and 200D vectors respectively, as we found it improved performance on the validation set). We find that our trained models generalize well, even when they are trained on Twitter or Ubuntu data and thus have absolutely no knowledge of the HUMOD domain. This is particularly impressive as in addition to the difference in domain there are also differences in pre-processing between the Twitter, Ubuntu, and HUMOD datasets.

Compared to the supervised baseline metrics reported by \citet{Merdivan_2020}, we perform significantly better than the second-best supervised baseline, HAN-R(MSE) with a reported correlation of 0.128, but worse than the supervised BERT baseline that achieved a reported correlation of $0.602$. However, the BERT metric was both fine-tuned to HUMOD and the predictions were trained in a supervised manner. Given that our model is more interpretable, trained without any human annotated data, and performs domain generalization, these results nevertheless demonstrate that using classic techniques in this novel way holds promise. Future work in this direction is merited. 

As a followup experiment, we train ULRoF on BERT features, dubbed ULRoF-BERT. To ensure a fair comparison with the \citet{Merdivan_2020} baseline we re-implemented their supervised BERT baseline -- a single-layer 5-way classifier on BERT features. We use ADAM with a batch size of $6$, and a learning rate of $0.00005$ when fine-tuning ($0.001$ otherwise). We use \texttt{bert-base-uncased} from HuggingFace \cite{wolf2019huggingfaces} as our pretrained BERT. We train all models for 20 epochs; we checkpoint after each epoch and test the checkpoint with the highest performance on our HUMOD validation set.

We train the baseline and ULRoF-BERT architectures both with and without fine tuning of the BERT model (i.e., updating the pretrained weights). To determine the generalization of the unsupervised models we train on HUMOD, Ubuntu, and Twitter data before determining the model correlation with the human annotations on the HUMOD test set (see Table \ref{tab:BERT_models}). We found poor performance using randomly samples as negative examples, and instead used ``I don't know'' as our negative examples. 

We immediately note that allowing the unsupervised model to fine-tune improves performance when training on HUMOD data, but causes catastrophic drops in performance when training on data from another domain. As expected, the full-capacity model overfits to the train domain. It is clear that we are overfitting to the training domain and not the specifics of the training set as we achieve excellent performance when the train and test data are both from the HUMOD domain. Next, we note ULRoF-BERT achieves good performance and much better generalization between domains. This demonstrates that even a simplistic model such as ULRoF can achieve good performance given the proper features. It also shows that our hand-engineered features vary less by domain than BERT features as ULRoF-BERT's drop in performance is two to five times worse than in either ULRoF-1 or ULRoF-2. Finally, the fact that BERT features are intended to be very general and that we achieve good performance without tuning them lends support to our proposal for the use of general linguistic features as input to a low-capacity model. What remains an open question is how scalably a hand-engineered set of features can be developed. Even if the hand-engineered approach proves to be impractical, our work suggests that less-interpretable but highly generic text features combined with unsupervised learning is another promising path to a dialogue evaluation metric that maintains the desiderata of not requiring human annotated data while avoiding overfitting to domain; albeit now without the benefit of interpretability.

\begin{table}[ht]
\resizebox{\textwidth}{!}{%
\begin{tabular}{llll}
                           & \multicolumn{3}{c}{Training Data Domain}  \\ \cline{2-4} 
\multicolumn{1}{c|}{Model} & \multicolumn{1}{c}{HU.} & \multicolumn{1}{c}{Ub.} & \multicolumn{1}{c|}{Tw.} \\ \hline
\multicolumn{1}{|l|}{Sup. BERT + FT} & 0.563 &  -  &  - \\
\multicolumn{1}{|l|}{Sup. BERT} & 0.627  &  - & - \\
\multicolumn{1}{|l|}{Unsup. BERT + FT} & 0.651 & 0.030 &  0.240 \\
\multicolumn{1}{|l|}{ULRoF-BERT} & 0.591 & 0.487 & 0.492 \\ \cline{1-1}
\end{tabular}%
}
\caption{Pearson's correlation between average human rating and model predictions on the HUMOD test set for relevance BERT models. Row is the relevance model with +FT used to indicate whether the BERT model was fine-tuned. Column is the domain of the training data. The architectures and loss of the last two rows are the same, the difference is whether BERT is fine-tuned. Note that by not fine-tuning BERT we achieve far better domain generalization.}
\label{tab:BERT_models}
\end{table}

\section{Conclusions}

In this work, we outline known shortcomings affecting existing dialogue metrics, and propose a feature-based approach to circumvent them. We demonstrate that feature-based approaches can be well-suited for zero-shot generalization between domains, both by an analysis of the features and by training a model on the related task of relevance prediction. By using a low-capacity model with well-chosen features, we can generalize well to unseen domains. Furthermore, this approach has the desirable property of being easily interpreted. At test time, the exact value, meaning, and definition of each feature is known, and with a model such as ULRoF we also know the relative importance of each feature, as encoded by regression weights. 

Our ULRoF experiments also show that training and evaluation can be done without gold-standard responses at test time, and without expensive human-annotated data. Although our ULRoF experiment is for relevance prediction, the triplet loss can be modified for dialogue evaluation by introducing additional types of negative responses, such as repeating the context in response, or jumbling the word order. We believe that the low capacity of the model, combined with generic features, should allow such an approach to avoid the overfitting problems that would impact a deep model trained in this manner.

Now that we have demonstrated that this approach is sound in principle, future work will include the determination of a full set of linguistic features appropriate for dialogue evaluation (particularly features regarding dialogue acts and dialogue breakdowns \cite{Breakdown1, Chinaei2017}), to extend our existing training procedure to train for overall response quality instead of relevance, and to obtain human response quality scores for dialogues from a variety of domains so that the final set of features can be fully evaluated and compared to existing methods.

\bibliography{tacl2018}
\bibliographystyle{acl_natbib}

\end{document}